\documentclass[11pt]{article}

\usepackage[final]{acl}

\usepackage{times}
\usepackage{latexsym}
\usepackage{booktabs}
\usepackage{comment}
\usepackage[T1]{fontenc}

\usepackage[utf8]{inputenc}

\usepackage{microtype}

\usepackage{inconsolata}

\usepackage{graphicx}
\usepackage{subcaption}
\usepackage{amsmath}
\usepackage{amssymb}
\usepackage{mathtools}
\usepackage{amsthm}
\DeclareMathOperator*{\Dir}{Dir}
\DeclareMathOperator*{\DP}{DP}

\usepackage{titlesec}

\title{An Information Bottleneck for Post-hoc Memorization Mitigation in LLMs}

\author{%
    Dina El Zein$^{1, 2}$ \quad Shashi Kumar$^{1, 2}$ \quad James Henderson$^{1}$ \\
    $^{1}$Idiap Research Institute, Switzerland \quad
    $^{2}$EPFL, Switzerland \\
    \texttt{\{dina.el-zein, shashi.kumar, james.henderson\}@idiap.ch}
}


\begin{document}

\maketitle

\begin{abstract}

Recent studies have shown that large language models (LLMs) can memorize and reproduce copyrighted training data, raising serious privacy and intellectual property concerns. In this work, we investigate whether memorization can be mitigated without modifying or retraining the underlying model through an information bottleneck. We propose a modular plug-and-play framework that inserts a nonparametric variational information bottleneck (NVIB) between a frozen LLM backbone and its language modeling head. By compressing the final hidden representations before token prediction, the proposed bottleneck aims to preserve general linguistic knowledge while filtering out highly specific memorized sequences. We evaluate our approach on memorization extraction tasks involving book passages and on standard language modeling benchmarks to measure the trade-off between memorization reduction and downstream utility. Our experiments show that the NVIB layer substantially reduces extractability while maintaining competitive performance on benchmark tasks. These results suggest that adding an information bottleneck offers a practical and architecture agnostic direction for post-hoc memorization mitigation in pretrained LLMs.

\end{abstract}

\section{Introduction}


Large Language Models (LLMs) are widely used across natural language processing tasks, including text generation, question answering, summarization, translation, reasoning, and multimodal applications~\cite{raffel2020exploring}. Trained on massive datasets collected from books, websites, and other textual sources~\cite{raffel2020exploring}, they learn statistical relationships that enable coherent text generation~\cite{cooper2025extracting, zhang2020dialogpt}. However, increasing model and dataset scales have amplified concerns about memorization of training data.


LLMs can retain and reproduce specific training sequences, particularly rare or distinctive passages~\cite{carlini2021extracting, carlini2022quantifying}, raising concerns about copyrighted or sensitive content. Recent studies demonstrate that open-weight LLMs can reproduce substantial portions of copyrighted books nearly verbatim~\cite{ahmed2026extracting}. Extraction attacks further reveal that even production-level LLMs equipped with safety mechanisms can generate long copyrighted passages with high fidelity, raising concerns about privacy, intellectual property, and responsible deployment.

Several approaches have been proposed to mitigate memorization in LLMs. Data deduplication and filtering techniques can reduce the occurrence of repeated training examples, but they require preprocessing before training and cannot fully eliminate memorization~\cite{lee2022deduplicating}. Other approaches, such as differential privacy or retraining-based mitigation strategies, often introduce significant computational costs and may negatively impact downstream model performance~\cite{abadi2016deep}. Thus, efficient post-hoc methods are needed to reduce memorization without modifying or retraining the underlying model.


In this paper, We propose a plug-and-play nonparametric variational information bottleneck (NVIB) framework~\cite{henderson2023a} for memorization mitigation in pretrained LLMs\footnote{The code will be made publicly available upon publication of the paper.}. Our approach inserts an NVIB layer between the frozen LLM backbone and its language modeling head, compressing latent representations before token prediction. The bottleneck aims to reduce verbatim memorization while preserving downstream performance. Prior work on NVIB-based latent information filtering has demonstrated promising results in privacy-preserving representation learning, maintaining utility while improving privacy guarantees~\cite{elzein2026differential, elzein2026nonparametric}.

Our contributions are summarized as follows:
\begin{itemize}\addtolength{\itemsep}{-1ex}
\item We introduce a plug-and-play NVIB bottleneck for post-hoc
memorization mitigation in pretrained LLMs.
\item We demonstrate reduced memorization extractability without
modifying or retraining the pretrained backbone.
\item We evaluate utility across knowledge, reasoning,
and instruction-following benchmarks, examining the trade-off
between memorization reduction and model performance.
\end{itemize}

\section{Background and Related Work}


Our proposed method builds on prior work in LLM memorization,
its measurement, and NVIB. We first review evidence that LLMs
reproduce training data, including copyrighted text. We then
discuss approaches for measuring memorization in generative
models. Finally, we introduce NVIB and its role in latent
compression and information filtering, which underpin our
memorization mitigation framework.

\subsection{Memorization in LLMs}

Memorization in LLMs occurs when a model reproduces portions of its training data during generation. More precisely, memorization occurs when a continuation $(s)$ generated from a prompt $(p)$ matches the corresponding sequence $([p|s])$ in the training dataset, where $([p|s])$  denotes the prompt $(p)$ followed by the generated continuation $(s)$~\cite{carlini2022quantifying}. Although LLMs learn general statistical and semantic patterns, they can also reproduce exact training sequences, particularly when these sequences are rare, unique, or highly distinctive~\cite{carlini2021extracting,carlini2022quantifying}.


LLM capabilities depend on the scale, diversity, and quality
of training data~\cite{su2026large}. Large text corpora allow
models to learn statistical relationships that support
performance across many NLP tasks~\cite{thrush2025improving}.
However, generalization can coexist with verbatim reproduction
of training examples~\cite{carlini2021extracting}.
Memorization can increase with model scale and dataset size
as models retain increasingly detailed training
information~\cite{carlini2022quantifying}, raising the risk
of extraction attacks and unintended disclosure.

Several studies have demonstrated that attackers can deliberately extract memorized data from LLMs using targeted prompting strategies~\cite{carlini2021extracting}. These attacks often rely on prefix prompting, sampling-based extraction, or repeated querying to induce the model to reproduce memorized sequences. Generated outputs can closely match training examples near-verbatim, revealing copyrighted, sensitive, or proprietary content~\cite{cooper2025extracting}. Recent work has further shown that even production-level language models equipped with safety mechanisms remain vulnerable to such extraction attacks, highlighting the persistence of memorization behavior in modern generative systems.

This phenomenon raises significant concerns regarding copyright protection, intellectual property rights, and responsible deployment of LLMs. Since many language models are trained on web-scale corpora that include copyrighted books, articles, and proprietary material, memorization may lead to unauthorized reproduction of protected content~\cite{xu2025copyright,cooper2025extracting}. Beyond copyright issues, memorization can result in leakage of sensitive information such as phone numbers, email addresses, or personal identifiers contained in the training data~\cite{luo2026llmsreallymemorizepersonally}. As a result, memorization has become an important challenge for the machine learning community and policymakers concerned with AI governance and data protection.

Existing memorization mitigation methods include dataset deduplication, differential privacy, retraining, and machine unlearning~\cite{lee2022deduplicating,abadi2016deep}. While these approaches have demonstrated varying degrees of success, they generally require preprocessing, expensive retraining, or utility trade-offs. In particular, deduplication requires access to the pretraining corpus and must be applied before training, making it unsuitable for models whose pretraining has been completed. These limitations motivate efficient post-hoc memorization mitigation methods that leave the pretrained model unchanged, as explored in this work.

\subsection{Measuring Memorization}


Measuring memorization is challenging because language models
can generalize while also retaining specific training
sequences~\cite{carlini2022quantifying}. Distinguishing
memorization from generalization therefore remains an
important research problem~\cite{wang2025generalization}.

One common approach for evaluating memorization relies on extraction-based methods, where generated model outputs are compared against known training samples~\cite{carlini2021extracting}. In these settings, adversaries provide carefully designed prompts or prefixes and analyze whether the model reproduces continuations that appear in the training corpus. Exact match and near verbatim generation are commonly used indicators of memorization, particularly when evaluating copyrighted books or sensitive textual data~\cite{cooper2025extracting}. Metrics such as extraction success rate, exact sequence recovery, and longest common substring overlap are frequently used to quantify the extent of memorized content leakage.

Another widely used approach is exposure analysis~\cite{carlini2019secret}, which measures how strongly a model has memorized a particular sequence relative to random chance. The underlying intuition is that memorized sequences are assigned disproportionately high likelihood compared to alternative candidate sequences. In practice, memorized samples often exhibit lower perplexity and higher probability scores than randomly sampled alternatives. Exposure-based metrics therefore provide a probabilistic measure of memorization that complements extraction-based evaluation.

Recent work has further demonstrated the effectiveness of scalable extraction attacks for recovering memorized training data from language models~\cite{nasr2025scalable}. These attacks use carefully designed prompts, repeated sampling, and decoding strategies to recover memorized sequences from model outputs. The results show that even large production-level language models can reproduce near verbatim training examples when queried appropriately, highlighting the practical risks associated with memorization in modern LLMs.

\subsection{Nonparametric Variational Information Bottleneck}\label{sec:nvib}

The NVIB~\cite{hendersonfehr2023} is a stochastic latent representation framework designed to regulate the flow of information through transformer attention layers. NVIB extends the variational information bottleneck (VIB) principle by introducing a latent mixture representation that compresses the input embeddings before they are processed by downstream attention computations. By constraining the amount of information retained in the latent space, NVIB encourages the model to preserve high-level semantic structure while filtering unnecessary or highly specific information.

\paragraph{Latent Mixture Representation}
Rather than directly operating on the original set of input embeddings, NVIB represents the latent space as a stochastic weighted mixture of latent vectors. Given an input sequence of embeddings, the model samples a set of latent vectors $\mathbf{Z}$ together with associated normalized weights $\boldsymbol{\pi}$. This latent representation is then consumed by a modified attention mechanism referred to as denoising attention~\cite{hendersonfehr2023}, where the weights $\boldsymbol{\pi}$ bias the attention weights.
To model variable length latent representations, NVIB employs Bayesian nonparametric methods based on a Dirichlet Process (DP). This allows the model to define probability distributions over potentially unbounded sets of weighted vectors without fixing the number of latent mixture components in advance. 

\paragraph{Prior}
The uninformative latent mixture representation is modeled using a Dirichlet Process prior, $F \sim \DP(G_0^p,\alpha_0^p)$, where $G_0^p$ denotes the base distribution and $\alpha_0^p$ controls the concentration of the mixture weights. The latent vectors are sampled from Gaussian components parameterized by a zero-mean unit-variance prior distribution $G_0^p$. The corresponding mixture weights follow a symmetric Dirichlet distribution parameterized by $\alpha_0^p$, encouraging sparse latent representations in which only a small number of latent vectors receive significant probability mass.

\paragraph{Posterior}
Given an input sequence of embeddings $\mathbf{x}\in\mathbb{R}^{n\times d}$, the NVIB layer predicts a set of posterior parameters consisting of pseudo-counts $\alpha_i^q \in \mathbb{R}$ together with Gaussian parameters $(\boldsymbol{\mu^q_i} \in \mathbb{R}^{d}, (\boldsymbol{\sigma^q_i})^2 \in \mathbb{R}^{d})$, one for each position $1\leq i\leq n$ plus one for the prior $i=n+1$. These parameters define a posterior Dirichlet Process distribution over the latent mixture representation:
\vspace{-0.5em}
{
\setlength{\abovedisplayskip}{4pt}
\setlength{\belowdisplayskip}{4pt}
\setlength{\abovedisplayshortskip}{2pt}
\setlength{\belowdisplayshortskip}{2pt}
\begin{align}
F &\sim \DP(G_0^q,\alpha_0^q), \qquad
\alpha_0^q &= \sum_{i=1}^{n+1}\alpha_i^q,
\end{align}
}
where the base distribution $G_0^q$ is represented as a weighted mixture of Gaussian components:
{\setlength{\abovedisplayskip}{4pt}
\setlength{\belowdisplayskip}{4pt}
\setlength{\abovedisplayshortskip}{2pt}
\setlength{\belowdisplayshortskip}{2pt}
\begin{align}
G_0^q = \sum_{i=1}^{n+1}\frac{\alpha_i^q}{\alpha_0^q}G_i^q,\qquad 
G^q_i &= \mathcal{N}(\boldsymbol{\mu^q_i},\boldsymbol{I}(\boldsymbol{\sigma^q_i})^2).
\end{align}
}
During training, latent vectors are sampled from the posterior distribution as
{\setlength{\abovedisplayskip}{4pt}
\setlength{\belowdisplayskip}{4pt}
\setlength{\abovedisplayshortskip}{2pt}
\setlength{\belowdisplayshortskip}{2pt}
\begin{align}
\boldsymbol{\pi}\sim\Dir(\boldsymbol{\alpha}^q),
\qquad
\boldsymbol{Z_i} \sim \mathcal{N}(\boldsymbol{\mu^q_i}, (\boldsymbol{\sigma^q_i})^{2}).
\end{align}
}

The stochastic latent representation acts as an information bottleneck by introducing controlled noise into both the latent vectors and their associated weights. Furthermore, the model can suppress individual latent components by assigning near zero pseudo-counts, effectively reducing the information capacity of the representation.

\paragraph{Training Objective}

The NVIB objective regularizes the amount of information transmitted through the latent representation. The loss combines the downstream task objective $L_T$ with KL-divergence regularization terms applied to both the Dirichlet mixture weights and the Gaussian latent variables:
\vspace{-0.5em}
\begin{equation}
L = L_T + \lambda_D L_D + \lambda_G L_G.
\end{equation}

The Gaussian regularization term $L_G$ encourages uncertainty in the latent vectors, while the Dirichlet regularization term $L_D$ promotes sparse latent mixtures by reducing the contribution of less informative components. Together, these regularizers constrain the information capacity of the latent space, encouraging compressed representations that preserve task-relevant semantic information while filtering fine-grained details that may contribute to memorization.

In this work, we insert the NVIB layer between the frozen LLM backbone and its language modeling head. By compressing the final hidden representations before token prediction, the proposed framework aims to reduce the recovery of memorized training sequences while preserving downstream language modeling capabilities. Prior work has investigated NVIB in pretrained transformer architectures, including its integration within pretrained models followed by fine-tuning ~\cite{fehr2024nonparametric,fehr2025finetuning}. Other studies have applied NVIB to encoder-based models such as BERT and RoBERTa, comparing it against alternative bottleneck mechanisms, including VIB, fixed Gaussian noise, and learned Gaussian noise, while also examining the effects
of its regularization components ~\cite{elzein2026differential,elzein2026nonparametric}.
Building on these findings, we investigate NVIB for post-hoc memorization mitigation in decoder-only LLMs. Unlike these approaches, we insert the bottleneck between the pretrained backbone and its language modeling head, keeping both frozen and optimizing only the added bottleneck parameters.

\begin{figure*}[t]
    \centering
    \includegraphics[width=0.95\linewidth]{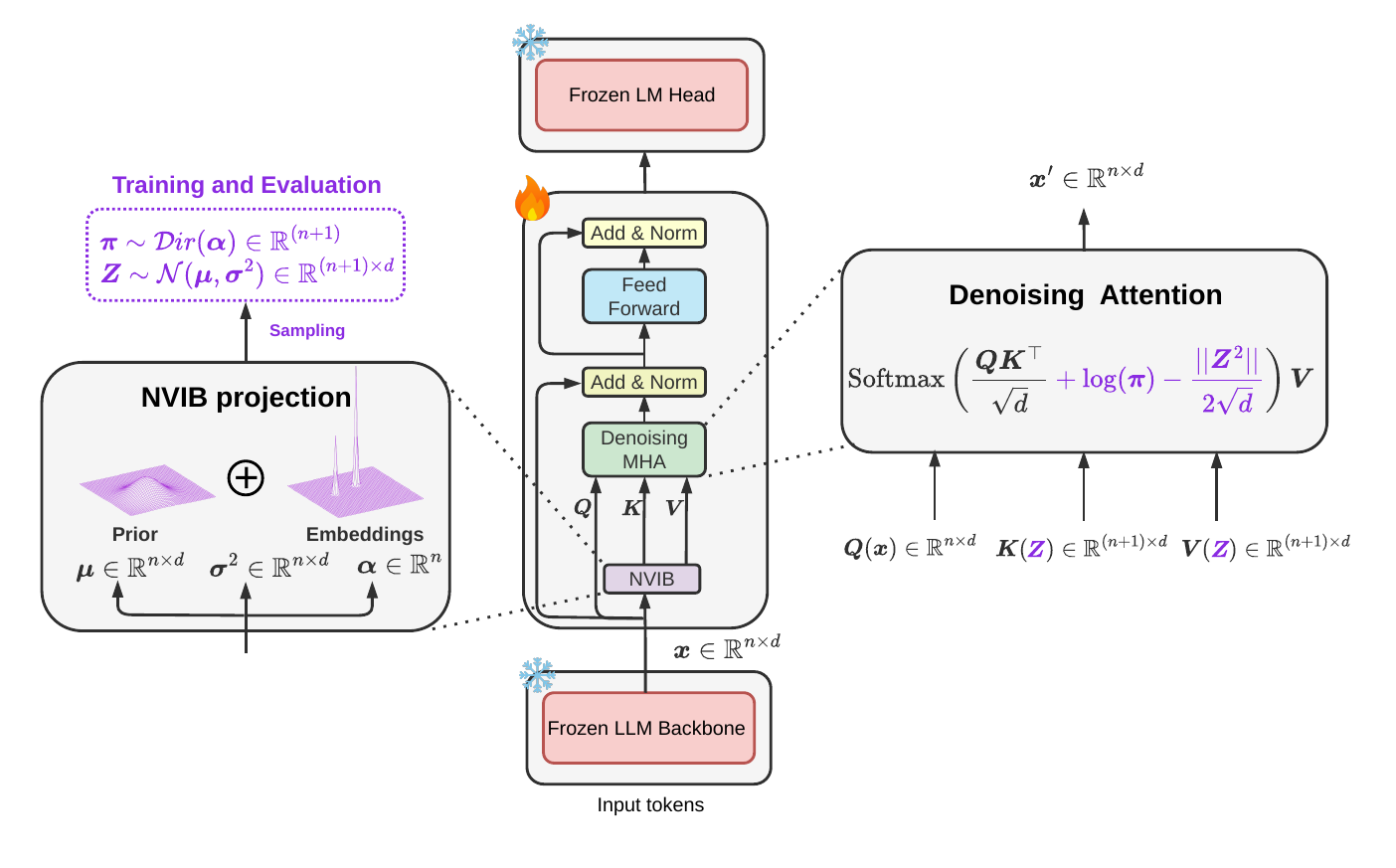}
    \caption{\textbf{Overview of the proposed memorization bottleneck architecture.} A frozen decoder-only LLM backbone generates the hidden representations $\mathbf{x}\in\mathbb{R}^{n\times d}$, which pass through a trainable NVIB block before the frozen language modeling (LM) head. The block combines a stochastic bottleneck, denoising attention, and feed-forward layers to reduce memorization while preserving language modeling performance. Only the bottleneck parameters are trained.}

    \label{fig:method}
\end{figure*}
\section{Memorization Bottleneck for Large Language Models}\label{sec:method}
In this work, we propose a memorization bottleneck for pretrained LLMs based on NVIB. The key intuition is that memorized training sequences rely on highly informative hidden representations that preserve fine-grained token-level information. By introducing a stochastic information bottleneck before token prediction, our approach compresses these representations to reduce the recoverability of memorized training data while preserving the semantic and linguistic information required for downstream language modeling.

\subsection{Memorization Bottleneck Architecture}

To mitigate memorization in LLMs, we integrate an NVIB bottleneck into frozen autoregressive LLMs such as Llama 3.1 and Qwen3. The bottleneck is inserted between the final hidden representations produced by the pretrained LLM backbone and the language modeling (LM) head used for token prediction. During training, only the parameters of the NVIB bottleneck are optimized, while the pretrained backbone and LM head remain frozen. This design provides an architecture-agnostic framework for post-hoc memorization mitigation.

Our approach builds on the NVIB~\cite{hendersonfehr2023}, which provides a stochastic latent representation framework for controlling the amount of information transmitted through transformer representations. We apply NVIB to the final hidden states of the LLM to compress the latent representation before token prediction.

The architecture of the proposed memorization bottleneck model is illustrated in Figure~\ref{fig:method}. Given an input sequence, the pretrained LLM produces a sequence of hidden representations denoted as $\mathbf{x} \in \mathbb{R}^{n \times d}$, where $n$ is the sequence length and $d$ is the hidden dimensionality. These hidden representations are processed by a lightweight transformer layer incorporating the NVIB mechanism. The NVIB layer projects each hidden representation into a set of posterior parameters consisting of pseudo-counts $\boldsymbol{\alpha^q}$ together with Gaussian parameters $(\boldsymbol{\mu^q}, (\boldsymbol{\sigma^q})^2)$. These parameters define a posterior Dirichlet Process distribution over sequences of weighted latent vectors.

More precisely, the latent representation is modeled as a stochastic sequence of weighted vectors $\mathbf{S}==(\boldsymbol{\pi},\mathbf{Z})$, sampled from the posterior distribution $Q(\mathbf{S})$. The latent weights $\boldsymbol{\pi}$ are sampled from a Dirichlet distribution, while the latent vectors $\mathbf{Z}$ are sampled from Gaussian components parameterized by the posterior mean and variance. Rather than interpreting the posterior as an exact Dirichlet Process realization, we interpret it as a stochastic sampling procedure that generates compressed latent representations of the original hidden states. The resulting latent sequence is then processed through a Denoising Multi-Head Attention (MHA) layer before being passed to the frozen LM head for token prediction. During training, latent vectors and mixture weights are sampled from their respective posterior distributions. At inference time, we use the posterior Gaussian means and normalized Dirichlet concentration parameters. 

The proposed stochastic bottleneck introduces controlled noise into the latent representation, reducing the model’s ability to preserve detailed token-level information. Since memorized training sequences often rely on high-fidelity hidden representations, this stochastic compression mechanism aims to suppress exact sequence retention while preserving broader semantic and linguistic structure necessary for downstream language modeling. Consequently, the bottleneck limits the recoverability of memorized training data while maintaining competitive utility on standard NLP tasks.

We adopt the NVIB formulation and regularization settings based on prior experimental findings~\cite{elzein2026differential,elzein2026nonparametric}.

\section{Experiments}
We evaluate the proposed memorization bottleneck along two dimensions: reducing the extraction of memorized copyrighted text and preserving downstream model utility. 

\subsection{Experimental Setup}

We evaluate the proposed memorization bottleneck framework on two open-weight LLM backbones: Llama-3.1-8B and Qwen3-8B. 

We compare NVIB against a Transformer-layer baseline that inserts a standard transformer layer at the same location. For both variants, only the added layer is trained, while the pretrained backbone and language modeling head remain frozen. Both variants are trained on the same data using a causal
language modeling task objective, while NVIB additionally
includes the regularization losses described in
Section~\ref{sec:nvib}. We train on the Open-R1 Mixture-of-Thoughts dataset~\cite{openr1,lozhkov2025openr1math220k}, which contains reasoning-oriented instruction-following data.

The NVIB bottleneck contains 251M trainable parameters for Llama-3.1-8B and 226M trainable parameters for Qwen3-8B, while the Transformer-layer baseline contains 218M and 193M, respectively. Full training and architectural configurations are provided in Appendix~\ref{app:training_details}.

\subsection{Memorization Evaluation}\label{sec:memorization_results}

To evaluate memorization, we adopt the probabilistic discoverable extraction metric of Hayes et al.~\cite{hayes2025measuring}, hereafter referred to as probabilistic extraction, following Cooper et al.~\cite{cooper2025extracting}. Unlike exact-match metrics, probabilistic extraction estimates the likelihood that a language model reproduces a target continuation from a given prefix.


Formally, let $z=z_{1:a+j}$ denote a sequence from the training corpus, where $z_{1:a}$ is a prefix of length $a$ and $z_{a+1:a+j}$ is the corresponding suffix of length $j$. Following Hayes et al.~\cite{hayes2025measuring}, the probabilistic extraction score of the sequence is defined as
\vspace{-0.5em}
{\setlength{\abovedisplayskip}{4pt}
\setlength{\belowdisplayskip}{4pt}
\setlength{\abovedisplayshortskip}{2pt}
\setlength{\belowdisplayshortskip}{2pt}
\begin{equation}
p_z \triangleq
\Pr_{\theta,\phi}
\left(
z_{a+1:a+j}\mid z_{1:a}
\right),
\label{eq:prob_extraction}
\end{equation}
}
where $\theta$ denotes the language model and $\phi$ denotes the decoding strategy. Since autoregressive language models factorize the joint probability over tokens, the extraction score can be written as
\vspace{-0.5em}
{\setlength{\abovedisplayskip}{4pt}
\setlength{\belowdisplayskip}{4pt}
\setlength{\abovedisplayshortskip}{2pt}
\setlength{\belowdisplayshortskip}{2pt}
\begin{align}
p_z&=\prod_{t=a+1}^{a+j}\Pr_{\theta,\phi}\left(z_t \mid z_{1:t-1}\right)
.
\label{eq:prob_extraction_factorization}
\end{align}
}

Intuitively, $p_z$ measures how likely the model is to generate the exact ground-truth continuation given the prefix. Larger values of $p_z$ indicate that the target continuation is more readily recoverable, whereas smaller values indicate reduced extractability. For each evaluated window, we compute a probabilistic extraction score $p_z$, and compare the distributions of these scores across model variants.

\begin{figure*}[!t]
\centering

\begin{subfigure}[t]{\textwidth}
\centering

\includegraphics[width=0.48\linewidth]{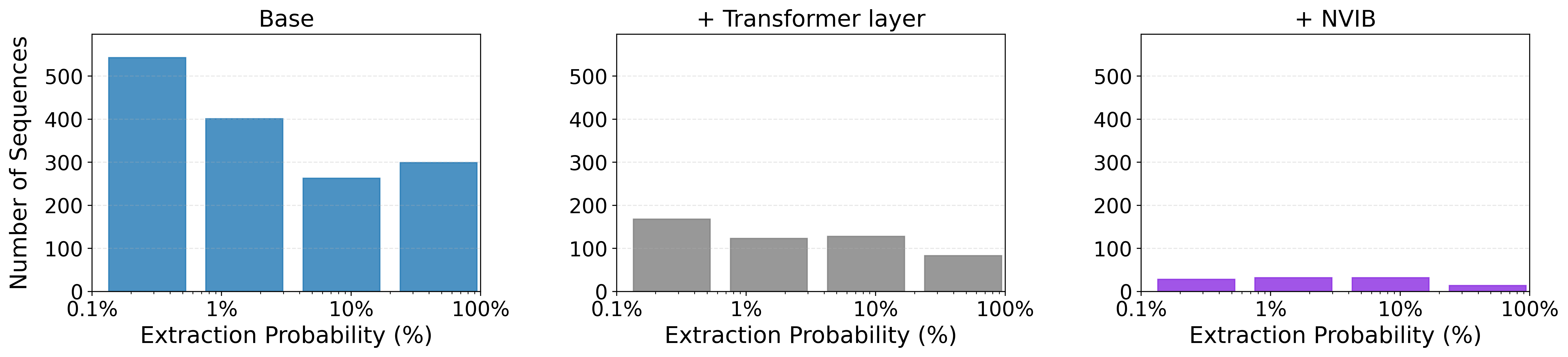}
\hfill
\includegraphics[width=0.48\linewidth]{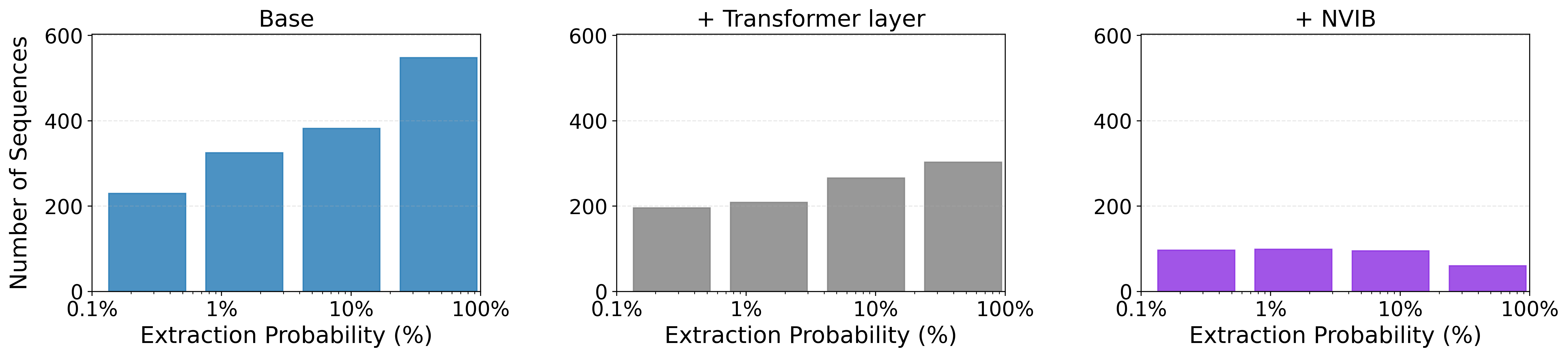}

\begin{minipage}{0.48\linewidth}
\centering
\small\textit{Animal Farm}
\end{minipage}
\hfill
\begin{minipage}{0.48\linewidth}
\centering
\small\textit{The Hitchhiker's Guide to the Galaxy}
\end{minipage}

\includegraphics[width=0.48\linewidth]{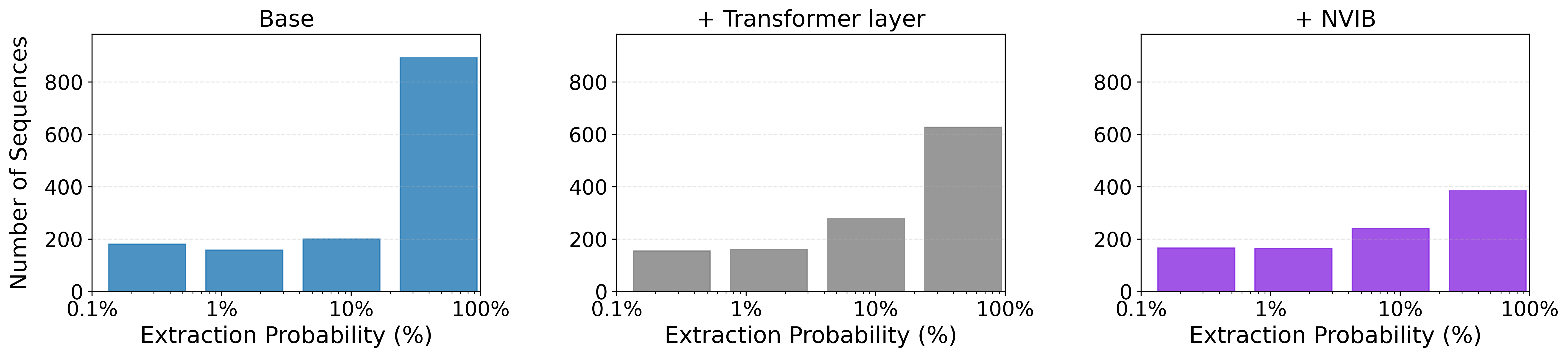}
\hfill
\includegraphics[width=0.48\linewidth]{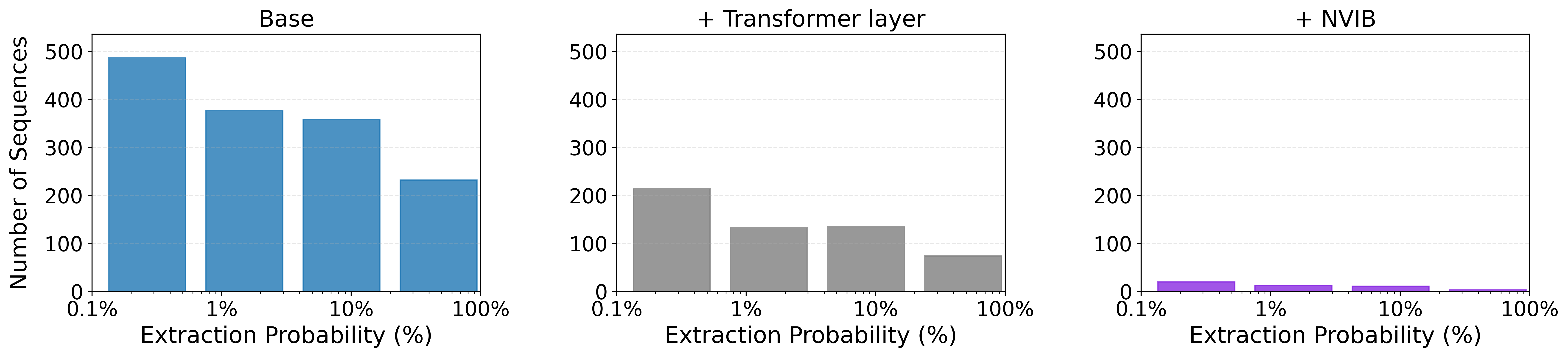}

\begin{minipage}{0.48\linewidth}
\centering
\small\textit{Crime and Punishment}
\end{minipage}
\hfill
\begin{minipage}{0.48\linewidth}
\centering
\small\textit{The Lord of the Rings: The Fellowship of the Ring}
\end{minipage}

\caption{\large\textbf{ Llama-3.1-8B}}
\label{fig:llama_books}

\end{subfigure}

\vspace{4mm}

\begin{subfigure}[t]{\textwidth}
\centering

\includegraphics[width=0.48\linewidth]{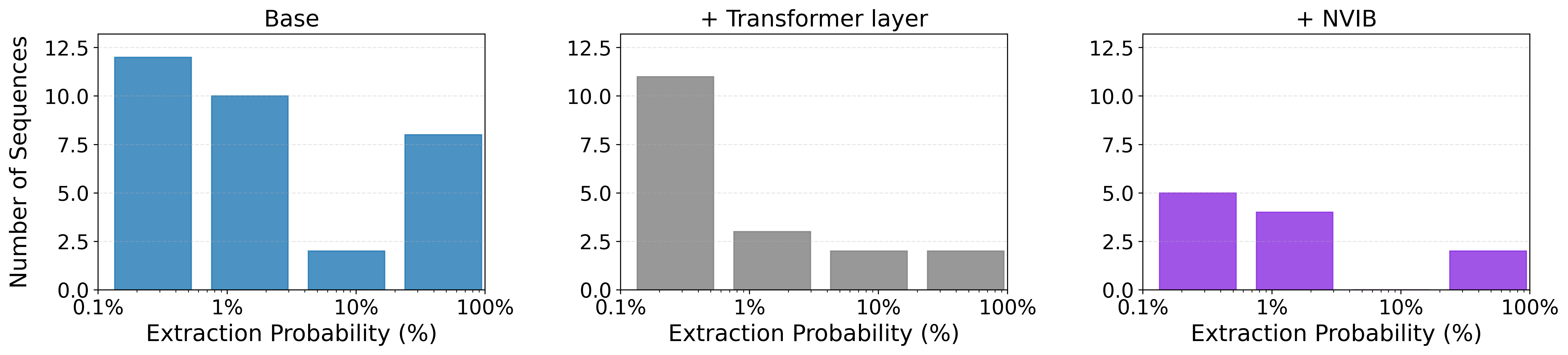}
\hfill
\includegraphics[width=0.48\linewidth]{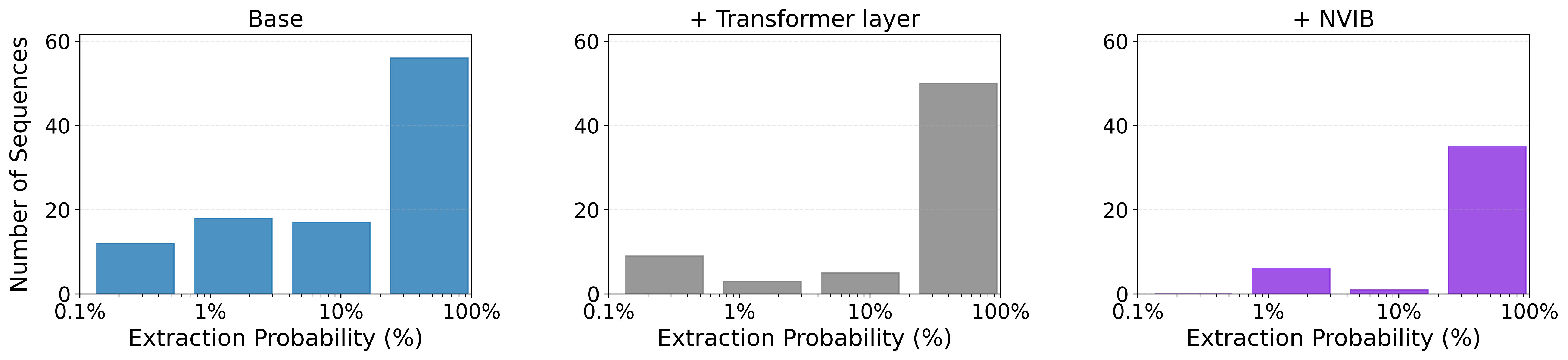}

\begin{minipage}{0.48\linewidth}
\centering
\small\textit{Animal Farm}
\end{minipage}
\hfill
\begin{minipage}{0.48\linewidth}
\centering
\small\textit{The Hitchhiker's Guide to the Galaxy}
\end{minipage}

\includegraphics[width=0.48\linewidth]{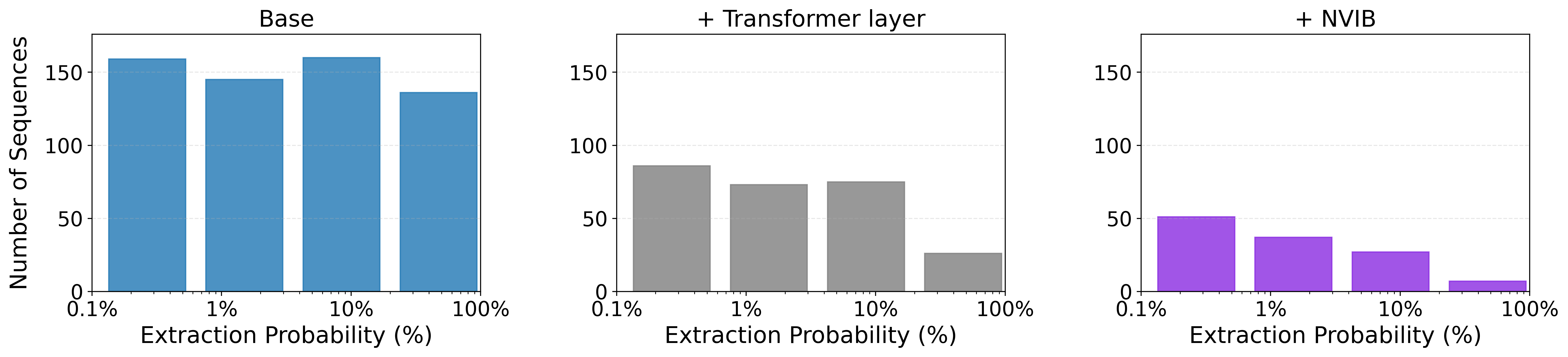}
\hfill
\includegraphics[width=0.48\linewidth]{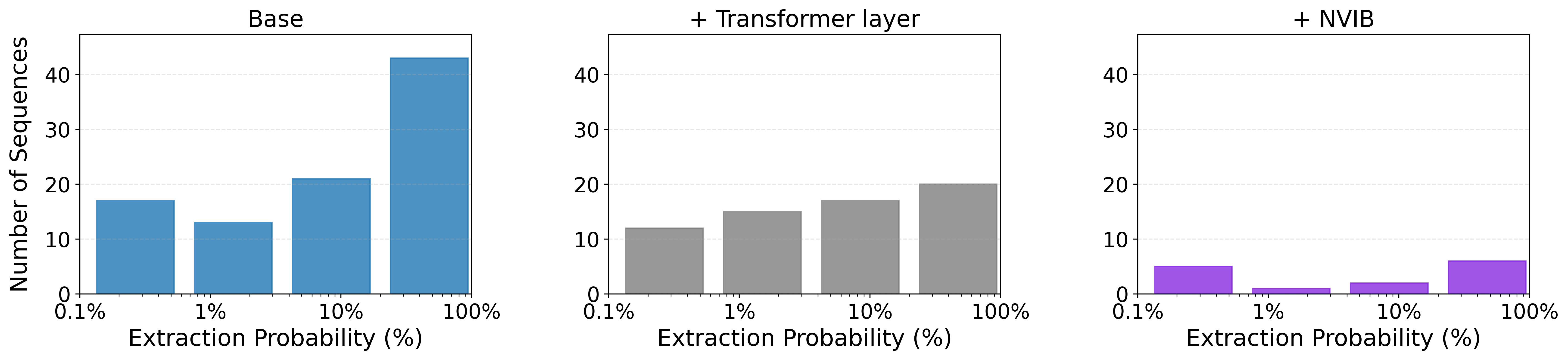}

\begin{minipage}{0.48\linewidth}
\centering
\small\textit{Crime and Punishment}
\end{minipage}
\hfill
\begin{minipage}{0.48\linewidth}
\centering
\small\textit{The Lord of the Rings: The Fellowship of the Ring}
\end{minipage}

\caption{\large\textbf{Qwen3-8B}}
\label{fig:qwen_books}

\end{subfigure}

\caption{\textbf{Distribution of probabilistic extraction scores on four copyrighted books.}
For each book, we compare the base model (Base), Transformer-layer baseline (+ Transformer layer), and NVIB-augmented model (+ NVIB). Histograms show extraction scores for sequences with $p_z \geq 10^{-3}$, grouped into four logarithmically spaced bins over $10^{-3} \leq p_z \leq 1$. The choice of cutoff is discussed in Appendix~\ref{app:threshold_sensitivity}. 
Lower extraction probabilities and fewer extracted sequences indicate reduced extractability.}

\label{fig:memorization}
\end{figure*}

We evaluate the original pretrained models (Base), Transformer-layer baselines (+ Transformer layer), and NVIB-augmented models (+ NVIB) on four copyrighted books: \textit{Animal Farm}, \textit{The Hitchhiker's Guide to the Galaxy}, \textit{Crime and Punishment}, and \textit{The Lord of the Rings: The Fellowship of the Ring}. Figure~\ref{fig:memorization} presents the extraction score distributions for Llama-3.1-8B and Qwen3-8B.

Following Cooper et al.~\cite{cooper2025extracting}, we compute probabilistic extraction scores for each model using their implementation and visualize the distribution of scores for all extracted sequences with $p_z \geq 0.001$. The extraction configuration determines the prefix and suffix lengths. Given an extraction window of length $L=100$ tokens, the first $L_p$ tokens form the prefix and the remaining $L_s$ tokens form the suffix, where $L_p + L_s = L$. For example, the \texttt{top40-suffix40} configuration uses top-$k$ decoding with $k=40$, a prefix length of $L_p=60$ tokens, and a suffix length of $L_s=40$ tokens. Probabilistic extraction is evaluated over overlapping 100-token windows with a stride of 10 tokens. 
The resulting histograms report the number of extracted sequences whose scores fall within each probability bin. Lower probabilistic extraction scores indicate that the model assigns lower probability to reproducing the target continuation, suggesting reduced extractability of memorized content. Consequently, a shift of the distribution toward lower extraction probabilities, together with a reduction in the number of extracted sequences, indicates that memorized passages are less readily recoverable.

\paragraph{Llama-3.1-8B.}

Figure~\ref{fig:llama_books} shows that NVIB substantially reduces extraction relative to both the base model and the Transformer-layer baseline across all four books. For \textit{Animal Farm}, the number of extracted sequences drops from 1,506 (Base) to 502 (+ Transformer layer) and just 106 (+ NVIB). Similar reductions are observed for \textit{The Fellowship of the Ring} (1,454 to 556 and 48), \textit{The Hitchhiker's Guide to the Galaxy} (1,485 to 974 and 351), and \textit{Crime and Punishment} (1,432 to 1,219 and 957), respectively. While the Transformer-layer baseline may reduce extraction by altering hidden representations passed to the frozen language modeling head, thereby disrupting memorized continuations, NVIB achieves substantially greater reductions. We hypothesize that this additional reduction arises from the information bottleneck suppressing fine-grained representations that support memorized token predictions. These results suggest that NVIB's effectiveness extends beyond adding a trainable layer, highlighting the contribution of the information bottleneck mechanism.

\begin{table*}[!t]

\centering
\scriptsize
\setlength{\tabcolsep}{3pt}
\renewcommand{\arraystretch}{1.1}
\begin{tabular}{lcccccc}
\toprule
& \multicolumn{3}{c}{\textbf{Qwen3-8B}}
& \multicolumn{3}{c}{\textbf{Llama-3.1-8B}} \\
\cmidrule(lr){2-4}
\cmidrule(lr){5-7}
\textbf{Book}
& \textbf{Base}
& \textbf{+ Transformer}
& \textbf{+ NVIB}
& \textbf{Base}
& \textbf{+ Transformer}
& \textbf{+ NVIB} \\
\midrule

Animal Farm
& $1.69 \times 10^{-4}$
& $4.90 \times 10^{-5}$
& $\mathbf{2.87 \times 10^{-5}}$
& $1.05 \times 10^{-2}$
& $2.85 \times 10^{-3}$
& $\mathbf{5.09 \times 10^{-4}}$ \\

Down and Out in the Magic Kingdom
& $3.61 \times 10^{-4}$
& $2.78 \times 10^{-4}$
& $\mathbf{8.12 \times 10^{-6}}$
& $7.58 \times 10^{-4}$
& $5.67 \times 10^{-4}$ 
& $\mathbf{4.42 \times 10^{-4}}$ \\

Crime and Punishment
& $6.60 \times 10^{-4}$
& $1.32 \times 10^{-4}$
& $\mathbf{6.39 \times 10^{-5}}$
& $4.96 \times 10^{-3}$
& $3.06 \times 10^{-3}$
& $\mathbf{1.67 \times 10^{-3}}$ \\

The Lord of the Rings: The Two Towers
& $5.90 \times 10^{-5}$
& $2.11 \times 10^{-5}$
& $\mathbf{7.54 \times 10^{-6}}$
& $9.54 \times 10^{-5}$
& $1.76 \times 10^{-5}$
& $\mathbf{4.00 \times 10^{-6}}$ \\

\bottomrule
\end{tabular}
\caption{\textbf{Mean probabilistic extraction scores.}
Scores are averaged over all evaluated windows, including those with $p_z < 10^{-3}$.
Lower is better; bold indicates the lowest score for each book and backbone.}
\label{tab:mean-extraction-main}
\end{table*}

\begin{table*}[!t]
\centering
\small
\setlength{\tabcolsep}{5pt}
\renewcommand{\arraystretch}{1.0}
\begin{tabular}{lcccccc}
\toprule
& \multicolumn{3}{c}{\textbf{Qwen3-8B}}
& \multicolumn{3}{c}{\textbf{Llama-3.1-8B}} \\
\cmidrule(lr){2-4}
\cmidrule(lr){5-7}
\textbf{Benchmark}
& \textbf{Base}
&\textbf{+ Transformer layer}
& \textbf{+ NVIB}
& \textbf{Base}
&\textbf{+ Transformer layer}
& \textbf{+ NVIB} \\
\midrule

\multicolumn{7}{c}{\textit{Knowledge \& Understanding}} \\
\midrule
MMLU & 49.7 & 48.2 & 49.5 & 64.0 & 60.0 & 59.8 \\
MMLU-Redux & 75.9 & 76.3 & 76.1 & 25.8 & 36.7 & 28.2 \\
\midrule
\multicolumn{7}{c}{\textit{Complex Reasoning}} \\
\midrule
MMLU-Pro & 63.6 & 62.2 & 62.4 & 36.2 & 32.3 & 32.3 \\
GPQA Diamond & 27.3 & 24.7 & 28.8 & 29.80 & 33.8 & 27.3 \\
MuSR         & 40.9 & 40.1 & 42.5 & 36.9 & 39.6 & 40.2 \\

\midrule
\multicolumn{7}{c}{\textit{Instruction Following}} \\
\midrule

IFEval       & 90.6 & 89.3 & 90.3 & 15.5 & 25.2 & 25.2 \\

\bottomrule
\end{tabular}
\caption{\textbf{Downstream utility across model variants.}
Performance of the original pretrained models (Base),
Transformer-layer baselines, and NVIB-augmented models
on knowledge and understanding, complex reasoning,
and instruction-following benchmarks.
Higher scores indicate better performance.}
\label{tab:quality_preservation_suite}
\end{table*}

\paragraph{Qwen3-8B.}

Figure~\ref{fig:qwen_books} shows a similar trend, with NVIB consistently reducing extraction beyond the Transformer-layer baseline. For \textit{Crime and Punishment}, extracted sequences decrease from 600 (Base) to 260 (+ Transformer layer) and 122 (+ NVIB). Similarly, extraction decreases from 94 to 64
and 14 for \textit{The Fellowship of the Ring}, from 103 to 67 and 42 for \textit{The Hitchhiker's Guide to the Galaxy}, and from 32 to 18 and 11 for \textit{Animal Farm},
following the same model variants. These results support NVIB's effectiveness across different model architectures, beyond the effect of adding a trainable layer. 
Additional probabilistic extraction results for Llama-3.1-8B and Qwen3-8B  are provided in Appendix~\ref{app:additional_extraction}, with decoding and prefix/suffix sensitivity analyses in Appendix~\ref{app:decoding_sensitivity}.


\paragraph{Mean Probabilistic Extraction Score.}
While the extraction histograms characterize sequences above the $p_z \geq 10^{-3}$ cutoff, they exclude windows with lower extraction probabilities to focus on the regions most relevant to memorization. For completeness, we also report the mean probabilistic extraction score $\overline{p}_z$, computed across all evaluated windows for each book and model variant, regardless of the cutoff. Table~\ref{tab:mean-extraction-main} shows that NVIB achieves the lowest mean score across all evaluated books for both Qwen3-8B and Llama-3.1-8B. These results confirm that NVIB's reduction in extraction probability extends across the full set of evaluated windows, while the thresholded histograms provide a more focused characterization of sequences with higher extraction probabilities. Further discussion of the cutoff and additional mean scores are provided in Appendices~\ref{app:threshold_sensitivity} and~\ref{app:extraction-probabilities}.

\subsection{Utility Evaluation}

To assess whether memorization mitigation compromises downstream
capabilities, we evaluate Qwen3-8B and Llama-3.1-8B across six
benchmarks covering three categories: (i) knowledge and
understanding, using MMLU~\citep{hendrycks2021measuring} and
MMLU-Redux~\citep{gema-etal-2025-done};
(ii) complex reasoning, using MMLU-Pro~\citep{wang2024mmlupro},
GPQA Diamond~\citep{rein2023gpqa}, and
MuSR~\citep{sprague2024musr}; and
(iii) instruction following, using
IFEval~\citep{zhou2023instruction}.
All evaluations are conducted using the
\texttt{lm-evaluation-harness} framework~\citep{eval-harness}.
We compare the original pretrained models (Base), the
Transformer-layer baseline, and the proposed NVIB-augmented
models to assess utility preservation.


Table~\ref{tab:quality_preservation_suite} presents the results.
For Qwen3-8B, NVIB largely preserves performance across all evaluated
capabilities, with only minor variations on knowledge, reasoning, and
instruction-following benchmarks. Notably, NVIB outperforms the
Transformer-layer baseline on five of the six benchmarks, suggesting
that the bottleneck preserves downstream capabilities beyond those
retained by a standard added layer.
For Llama-3.1-8B, utility preservation is more task-dependent. NVIB
exhibits moderate degradation on MMLU, MMLU-Pro, and GPQA Diamond,
while improving performance on MMLU-Redux and MuSR. The largest
improvement is observed on IFEval, with a gain of 9.7 percentage
points over the base model, although the Transformer-layer baseline
achieves the same performance. These findings highlight that the
utility trade-off depends on the underlying architecture and
evaluated capability, with some improvements attributable to the
additional trainable layer rather than the bottleneck itself.



Overall, the results suggest that NVIB can substantially
reduce memorization extractability while retaining much
of the evaluated downstream utility. The trade-off is
particularly favorable for Qwen3-8B, whereas Llama-3.1-8B
exhibits more pronounced differences across tasks.
Together with the memorization results, these findings
highlight the potential of post-hoc information
bottlenecks while emphasizing the importance of
evaluating capability-specific utility trade-offs.

\section{Conclusion}


We present a plug-and-play memorization mitigation framework for LLMs based on the Nonparametric Variational Information Bottleneck (NVIB). Our method inserts a stochastic bottleneck between the frozen LLM backbone and its language modeling head, optimizing only the bottleneck parameters. Experiments on Llama-3.1-8B and Qwen3-8B show that NVIB lowers probabilistic extraction scores and consistently reduces extractable passages from copyrighted books while preserving downstream performance. Compared with a standard Transformer-layer baseline, NVIB achieves greater reductions in extractability, suggesting benefits beyond adding a trainable layer. These results establish NVIB as a practical, architecture-agnostic approach to post-hoc memorization mitigation.

\section{Limitations}
Our study focuses on two model families and a selected set of copyrighted books, and future work could extend the evaluation to a broader range of architectures, datasets, and decoding settings. While the proposed bottleneck consistently reduces the extractability of memorized content, its effect may depend on the chosen compression strength and evaluation setup. Additional experiments could also examine other forms of sensitive information leakage beyond copyrighted text. Since the pretrained backbone remains frozen, our evaluation demonstrates reduced extractability of memorized content rather than removal of memorized information from the model parameters.


\bibliography{main}

\appendix


\section{Additional Memorization Evaluation}
\label{app:additional_memorization}

This appendix presents additional probabilistic extraction results and robustness checks that complement the representative examples discussed in Section~\ref{sec:memorization_results}.

\subsection{Threshold Sensitivity}
\label{app:threshold_sensitivity}


We assess the sensitivity of the probabilistic extraction histograms to the $p_z \geq 10^{-3}$ cutoff. Extraction scores are computed for all evaluated windows, but the cutoff determines which windows are included in the histograms. With the cutoff, only windows satisfying $p_z \geq 10^{-3}$ are displayed, whereas removing it includes windows with lower extraction scores as well. The histograms report raw counts within each probability bin rather than normalized proportions. Since windows below the cutoff have negligible extraction probabilities, including them would primarily increase the counts in the lowest probability bins, obscuring the differences between models in the higher-probability regions most relevant to memorization.

\subsection{Decoding and prefix-length sensitivity.}\label{app:decoding_sensitivity}
We assess decoding sensitivity using \textit{Harry Potter and the Half-Blood Prince} with Llama-3.1-8B. The numbers of extracted sequences for the base, Transformer-layer baseline, and NVIB-augmented models are
15 (Base), 2 (+ Transformer layer), and 1 (+ NVIB) ($T=1,k=40,L_p=50$), 
65 (Base), 22 (+ Transformer layer), and 18 (+ NVIB) ($T=1,k=40,L_p=60$),
320 (Base), 187 (+ Transformer layer), and 163 (+ NVIB) ($T=1,k=40,L_p=80$), and
2 (Base), 1 (+ Transformer layer), and 0 (NVIB) ($T=1,k=1,L_p=50$).
NVIB reduces the number of extracted sequences across all evaluated configurations, including greedy decoding ($k=1$).

For Qwen3-8B on \textit{Animal Farm}, NVIB also reduces probabilistic extraction under different prefix lengths. With $T=1$ and $k=40$, the numbers of extracted sequences are 32 (Base), 18 (+ Transformer layer), and 11 (+ NVIB) for $L_p=80$, and 1 (Base), 1 (+ Transformer layer), and 0 (+ NVIB) for $L_p=60$.

\subsection{Additional Probabilistic Extraction Results}\label{app:additional_extraction}

\begin{figure*}[p]
\centering

\begin{subfigure}[t]{0.48\textwidth}
    \centering
    \includegraphics[width=\linewidth]{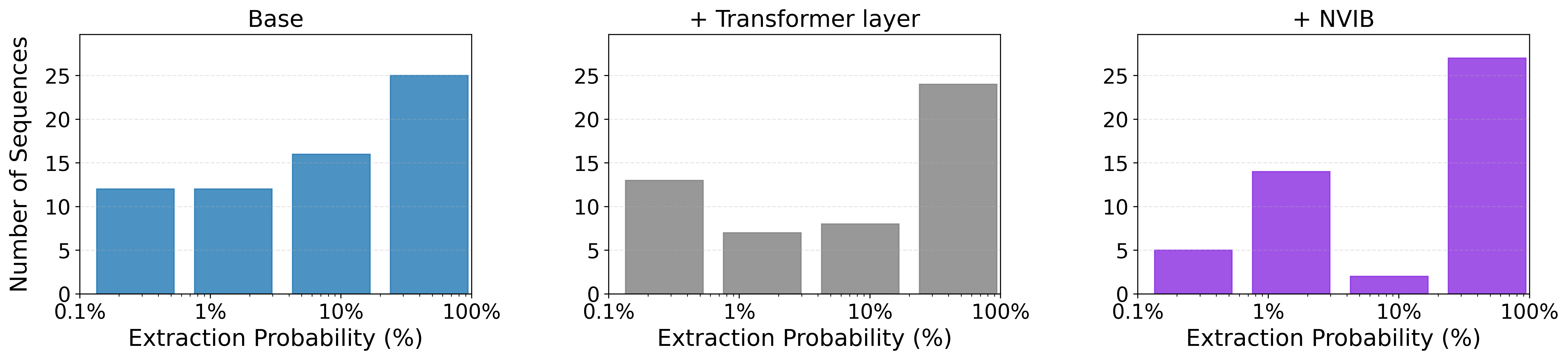}
    \subcaption{\textit{Down and Out in the Magic Kingdom} (Base: 65, + Transformer layer: 52, + NVIB: 48)}
\end{subfigure}
\hfill
\begin{subfigure}[t]{0.48\textwidth}
    \centering
   \includegraphics[width=\linewidth]{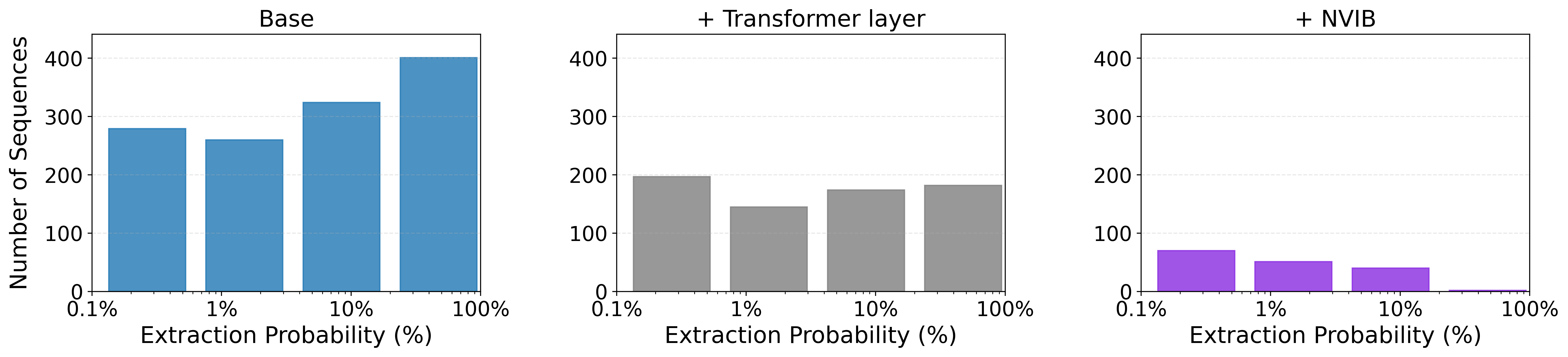}
    \subcaption{\textit{Harry Potter and the Sorcerer's Stone} (Base: 1264, + Transformer layer: 698, + NVIB: 163)}
\end{subfigure} 

\vspace{4mm}

\begin{subfigure}[t]{0.48\textwidth}
    \centering
    \includegraphics[width=\linewidth]{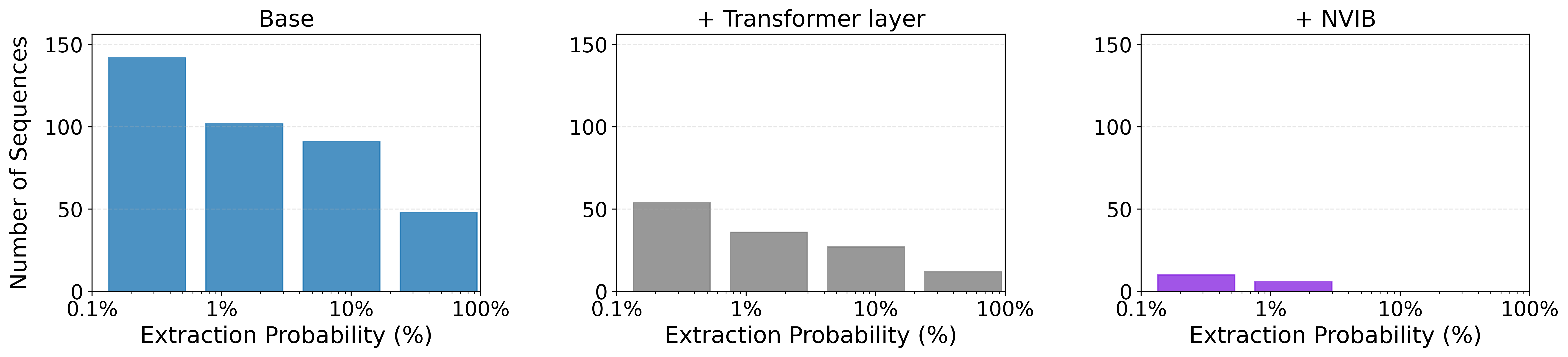}
     \subcaption{\textit{Harry Potter and the Chamber of Secrets} (Base: 383, + Transformer layer: 129, + NVIB: 16)}
\end{subfigure}
\hfill
\begin{subfigure}[t]{0.48\textwidth}
    \centering
    \includegraphics[width=\linewidth]{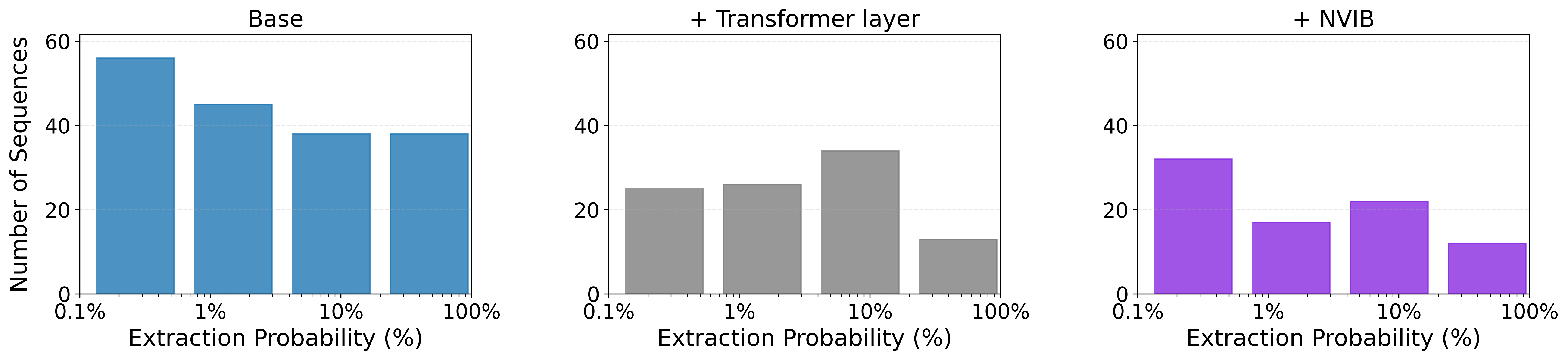}
    \subcaption{\textit{Harry Potter and the Prisoner of Azkaban} (Base: 177, + Transformer layer: 98, + NVIB: 83)}
    
\end{subfigure}

\vspace{4mm}

\begin{subfigure}[t]{0.48\textwidth}
    \centering
    \includegraphics[width=\linewidth]{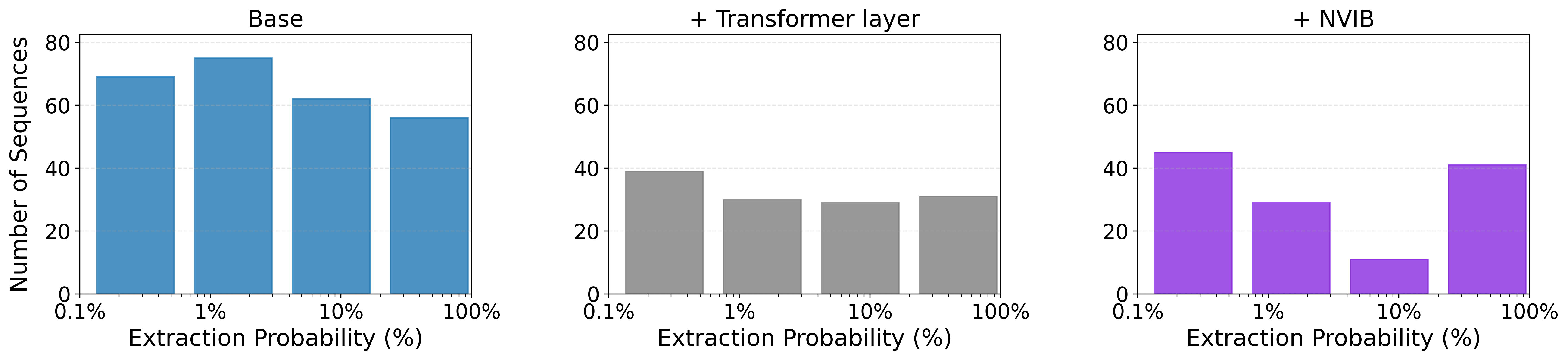}
    \subcaption{\textit{Harry Potter and the Goblet of Fire} (Base: 262, + Transformer layer: 129, + NVIB: 126)}
\end{subfigure}
\hfill
\begin{subfigure}[t]{0.48\textwidth}
    \centering
    \includegraphics[width=\linewidth]{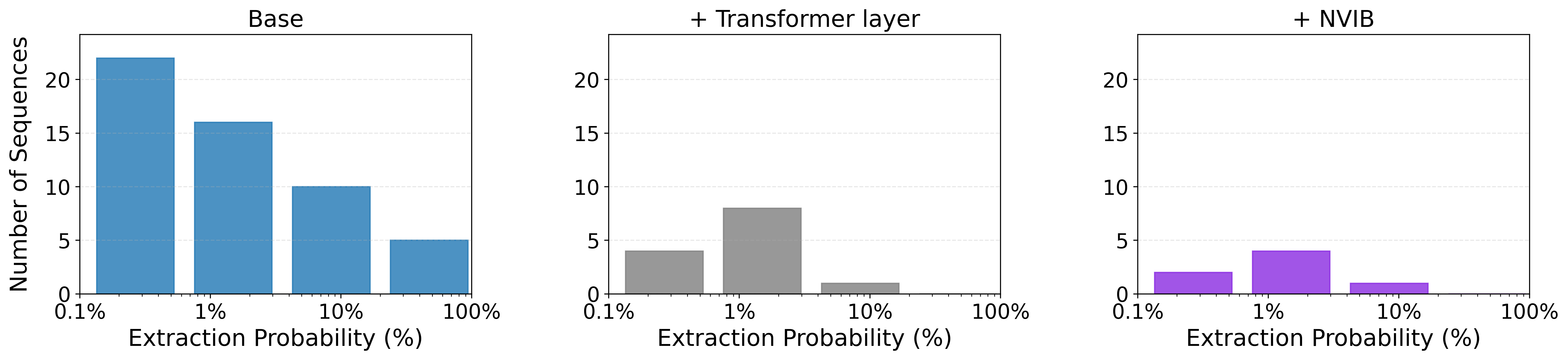}
    \subcaption{\textit{Harry Potter and the Order of the Phoenix} (Base: 53, + Transformer layer: 13, + NVIB: 7)}
\end{subfigure}

\vspace{4mm}

\begin{subfigure}[t]{0.48\textwidth}
    \centering
    \includegraphics[width=\linewidth]{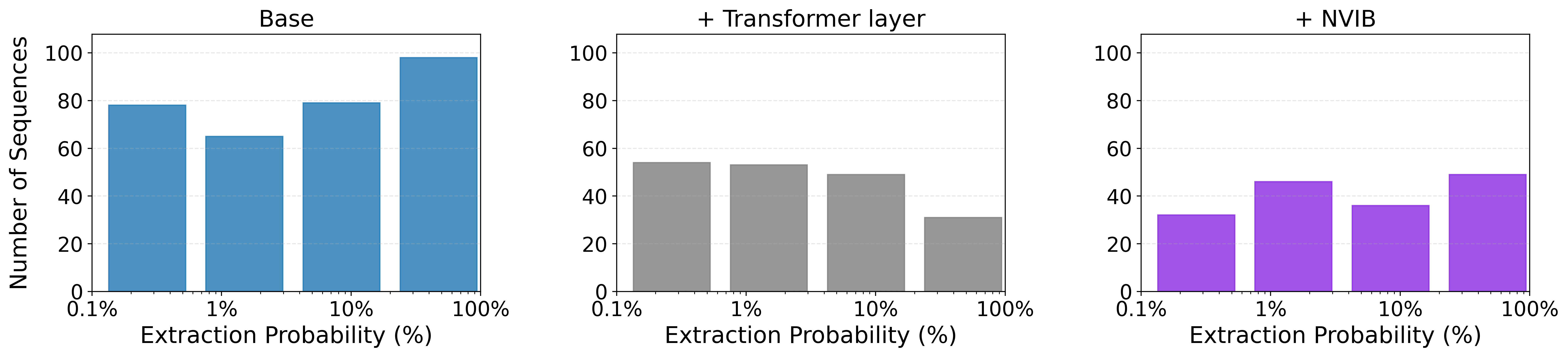}
    \subcaption{\textit{Harry Potter and the Half-Blood Prince} (Base: 320, + Transformer layer: 187, + NVIB: 163)}
\end{subfigure}
\hfill
\begin{subfigure}[t]{0.48\textwidth}
    \centering
    \includegraphics[width=\linewidth]{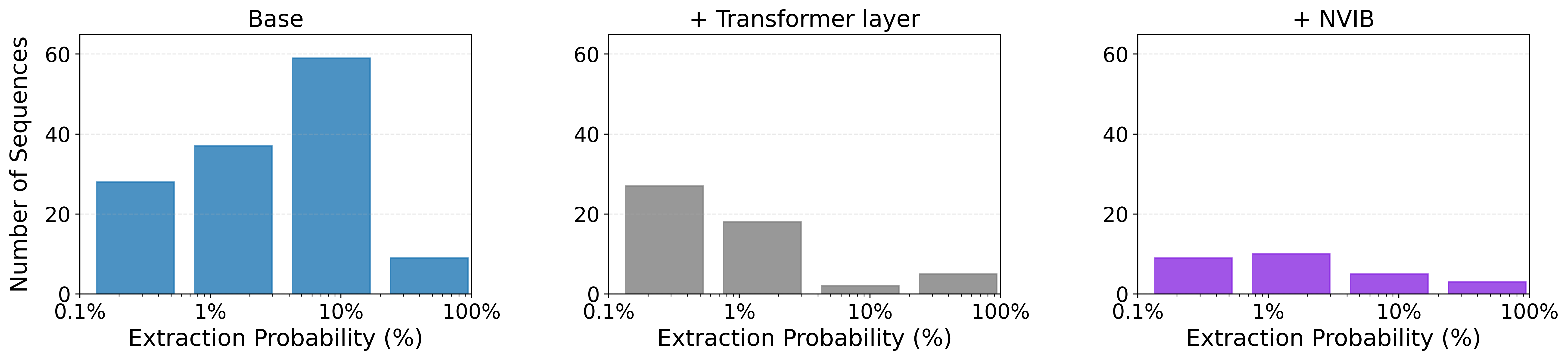}
    \subcaption{\textit{Harry Potter and the Deathly Hallows} (Base: 133, + Transformer layer: 52, + NVIB: 27)}
\end{subfigure}

\vspace{4mm}

\begin{subfigure}[t]{0.48\textwidth}
    \centering
    \includegraphics[width=\linewidth]{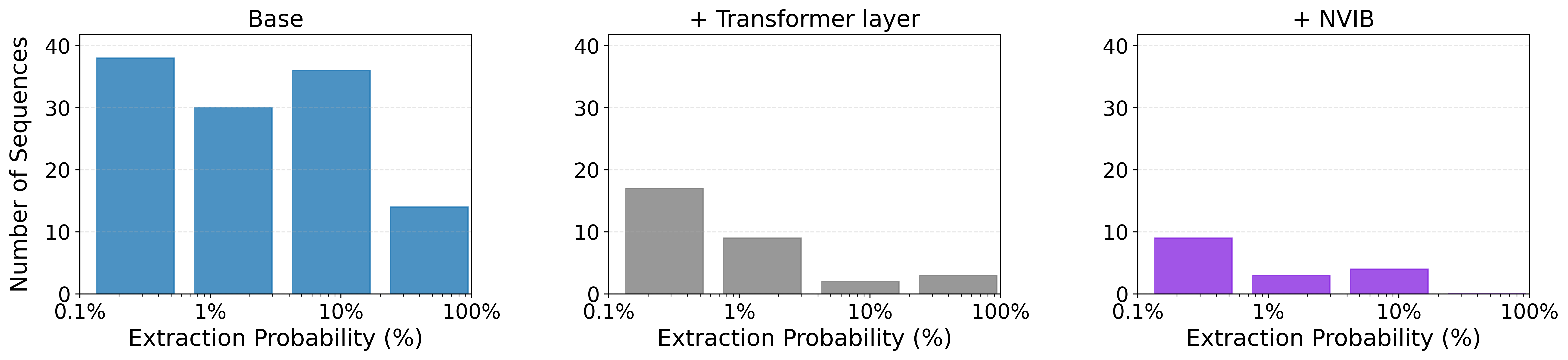}
    \subcaption{\textit{The Lord of the Rings: The Two Towers} (Base: 118, + Transformer layer: 31, + NVIB: 16)}
\end{subfigure}
\hfill
\begin{subfigure}[t]{0.48\textwidth}
    \centering
    \includegraphics[width=\linewidth]{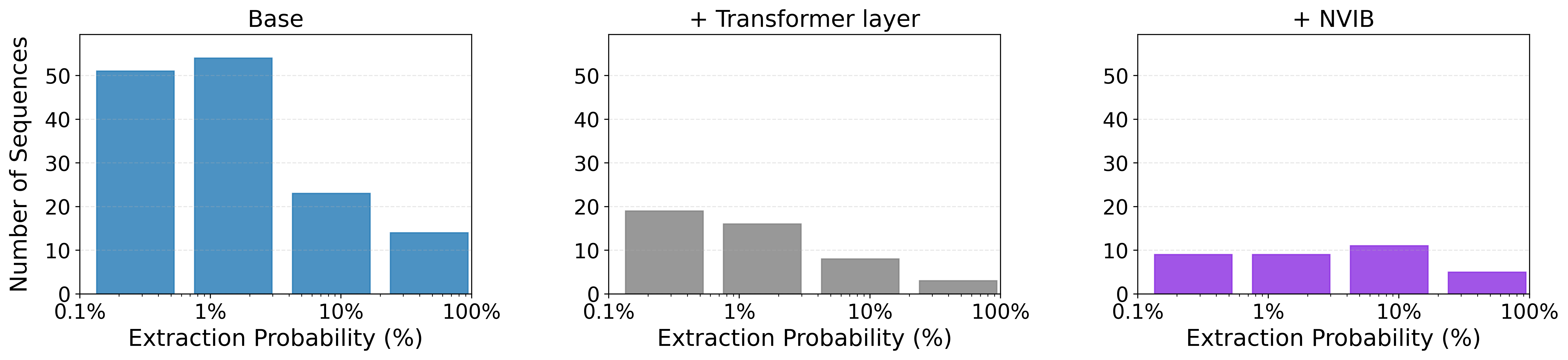}
    \subcaption{\textit{The Lord of the Rings: The Return of the King} (Base: 142, + Transformer layer: 46, + NVIB: 34)}
\end{subfigure}


\caption{
\textbf{Additional probabilistic extraction results for Llama-3.1-8B
on ten copyrighted books.} Each subfigure compares the base
model, the Transformer-layer baseline, and the NVIB-augmented
model. Histograms show the distributions of probabilistic
extraction scores for extracted sequences satisfying
$p_z \geq 0.001$. The reported counts indicate the number
of extracted sequences for each model.
}

\label{fig:appendix_llama_books}
\end{figure*}

\begin{figure*}[tp]
\centering
\begin{subfigure}[t]{0.48\textwidth}
    \centering
   \includegraphics[width=\linewidth]{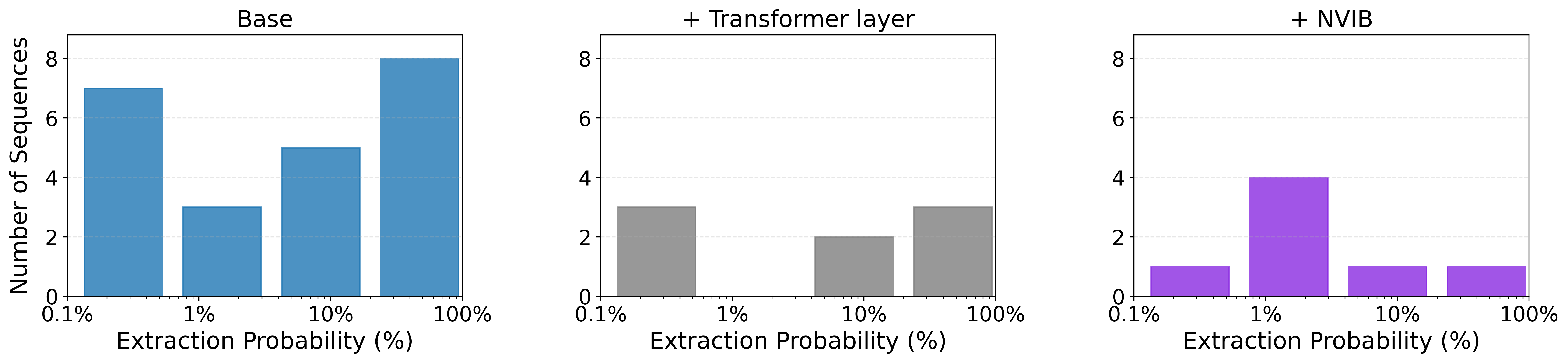}
     \subcaption{\textit{The Lord of the Rings: The Two Towers} (Base: 23, + Transformer layer: 8, + NVIB: 7)}
\end{subfigure}
\hfill
\begin{subfigure}[t]{0.48\textwidth}
    \centering
    \includegraphics[width=\linewidth]{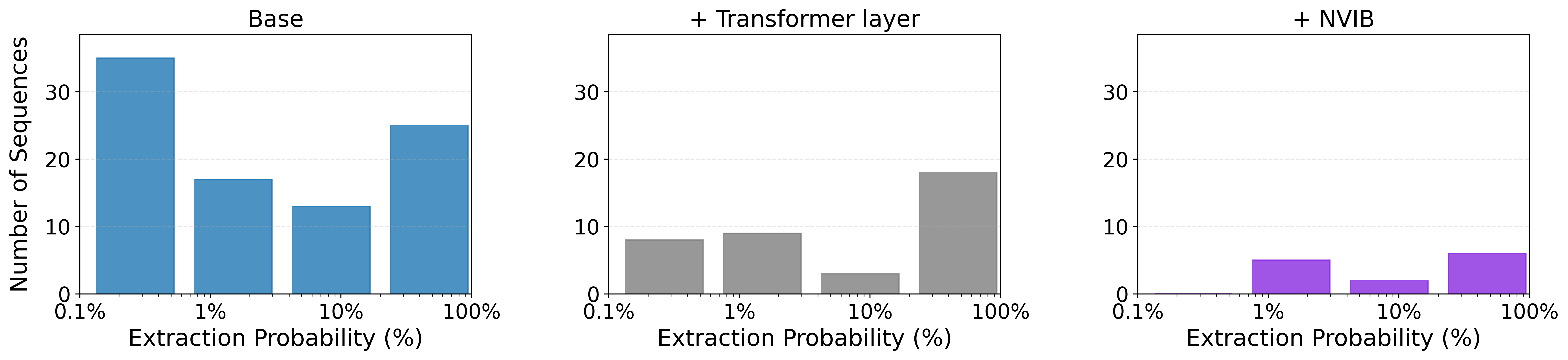}
   \subcaption{\textit{The Lord of the Rings: The Return of the King} (Base: 90, + Transformer layer: 38, + NVIB: 13)}
\end{subfigure} 

\vspace{4mm}

\begin{subfigure}[t]{0.48\textwidth}
    \centering
   \includegraphics[width=\linewidth]{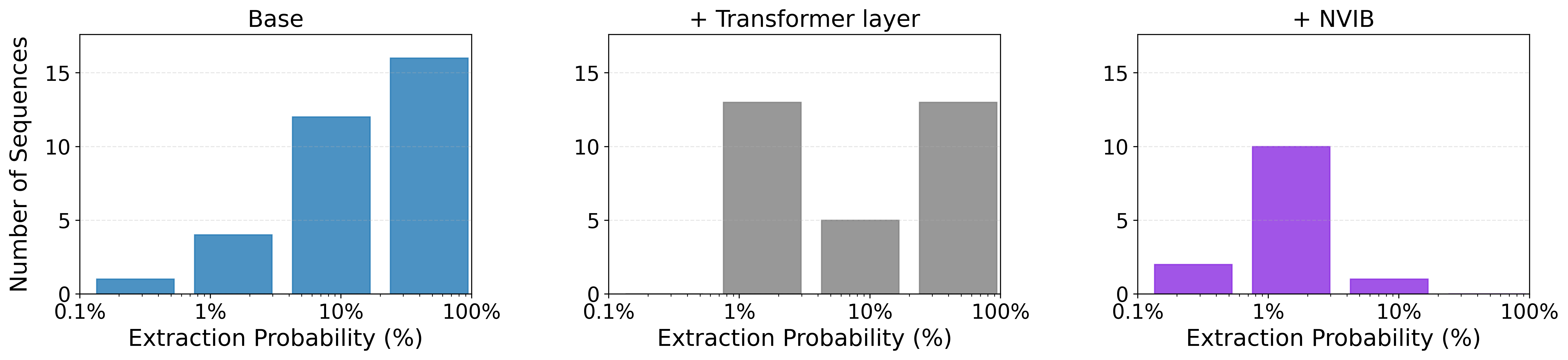}
     \subcaption{\textit{Down and Out in the Magic Kingdom} (Base: 33, + Transformer layer: 31, + NVIB: 13)}
\end{subfigure}

\caption{
\textbf{Additional probabilistic extraction results for Qwen3-8B
on three copyrighted books.} Each subfigure compares the
base model, the Transformer-layer baseline, and the
NVIB-augmented model. Histograms show the distributions
of probabilistic extraction scores for extracted sequences
satisfying $p_z \geq 0.001$. The reported counts indicate
the number of extracted sequences for each model.
}

\label{fig:appendix_qwen_books}
\end{figure*}

Figures~\ref{fig:appendix_llama_books} and~\ref{fig:appendix_qwen_books} present additional probabilistic extraction results for Llama-3.1-8B and Qwen3-8B, respectively, across ten and three copyrighted books. Each figure compares the base model, the Transformer-layer baseline, and the NVIB-augmented model.

Across all evaluated books, NVIB yields fewer extracted sequences than both the base model and the Transformer-layer baseline. For \textit{Down and Out in the Magic Kingdom} with Qwen3-8B, NVIB also shifts the distribution toward lower extraction probabilities, with fewer sequences in the high-probability bins. These results further support the effectiveness of NVIB in reducing probabilistic extraction across different copyrighted texts.

\captionsetup[subfigure]{labelformat=empty}

\subsection{Mean Probabilistic Extraction Scores}
\label{app:extraction-probabilities}

To complement the threshold-based extraction counts,
we report the mean probabilistic extraction score
for each book and model variant.
For a given book, let $\{z_i\}_{i=1}^{N}$
denote the evaluated windows, where $p_{z_i}$
is the extraction score of the $i$-th window,
as defined in Equation~\ref{eq:prob_extraction}.
The mean score is computed as
\begin{equation}
\overline{p}_z =
\frac{1}{N}\sum_{i=1}^{N}p_{z_i}.
\end{equation}

Unlike the threshold-based extraction counts,
the mean score summarizes the probabilities
across all evaluated windows without applying
the $p_z \geq 10^{-3}$ threshold.

Tables~\ref{tab:appendix_mean_extraction_llama}
and~\ref{tab:appendix_mean_extraction_qwen}
present the corresponding results for
Llama-3.1-8B and Qwen3-8B, respectively.

\begin{table*}[t]
\centering
\small
\setlength{\tabcolsep}{5pt}
\renewcommand{\arraystretch}{1.0}
\begin{tabular}{lccc}
\toprule
\textbf{Book}
& \textbf{Base}
& \textbf{+ Transformer}
& \textbf{+ NVIB} \\
\midrule

Harry Potter and the Sorcerer's Stone
& $4.88 \times 10^{-3}$
& $2.18 \times 10^{-3}$
& $\mathbf{8.61 \times 10^{-5}}$ \\

Harry Potter and the Chamber of Secrets
& $5.44 \times 10^{-4}$
& $1.64 \times 10^{-4}$
& $\mathbf{2.23 \times 10^{-6}}$ \\

Harry Potter and the Prisoner of Azkaban
& $3.07 \times 10^{-4}$
& $1.43 \times 10^{-4}$
& $\mathbf{9.74 \times 10^{-5}}$ \\


Harry Potter and the Order of the Phoenix
& $1.52 \times 10^{-5}$
& $1.25 \times 10^{-6}$
& $\mathbf{7.74 \times 10^{-7}}$ \\


Harry Potter and the Deathly Hallows
& $9.00 \times 10^{-5}$
& $2.26 \times 10^{-5}$
& $\mathbf{1.32 \times 10^{-5}}$ \\

The Lord of the Rings: The Fellowship of the Ring
& $1.27 \times 10^{-3}$
& $4.00 \times 10^{-4}$
& $\mathbf{2.30 \times 10^{-5}}$ \\


The Hitchhiker's Guide to the Galaxy
& $7.44 \times 10^{-3}$
& $4.10 \times 10^{-3}$
& $\mathbf{9.51 \times 10^{-4}}$ \\

\bottomrule
\end{tabular}
\caption{\textbf{Mean probabilistic extraction scores for
Llama-3.1-8B on additional copyrighted books.}
Lower values indicate lower extractability under the
evaluated decoding configuration.
Bold indicates the lowest mean score for each book.}
\label{tab:appendix_mean_extraction_llama}
\end{table*}

\begin{table*}[t]
\centering
\small
\setlength{\tabcolsep}{5pt}
\renewcommand{\arraystretch}{1.0}
\begin{tabular}{lccc}
\toprule
\textbf{Book}
& \textbf{Base}
& \textbf{+ Transformer}
& \textbf{+ NVIB} \\
\midrule

The Lord of the Rings: The Fellowship of the Ring
& $3.03 \times 10^{-4}$
& $1.58 \times 10^{-4}$
& $\mathbf{4.13 \times 10^{-5}}$ \\

The Lord of the Rings: The Return of the King
& $1.52 \times 10^{-4}$
& $1.06 \times 10^{-4}$
& $\mathbf{3.26 \times 10^{-5}}$ \\

The Hitchhiker's Guide to the Galaxy
& $1.24 \times 10^{-3}$
& $9.07 \times 10^{-4}$
& $\mathbf{5.43 \times 10^{-4}}$ \\

\bottomrule
\end{tabular}
\caption{\textbf{Mean probabilistic extraction scores for
Qwen3-8B on additional copyrighted books.}
Lower values indicate lower extractability under the
evaluated decoding configuration.
Bold indicates the lowest mean score for each book.}
\label{tab:appendix_mean_extraction_qwen}
\end{table*}

\section{Training Details}
\label{app:training_details}

We train both the NVIB bottleneck and the Transformer-layer baseline for 5 epochs, keeping the backbone model and LM head frozen. Both approaches use an effective batch size of 16 and a maximum sequence length of 8192 tokens. We use AdamW with an initial learning rate of $10^{-4}$, cosine learning-rate scheduling with 50 warmup steps, weight decay of 0.01, and bfloat16 precision.

For NVIB, we set $\lambda_D=\lambda_G=10^{-3}$, $\alpha_\tau=-10$, $\mu_\tau=1.0$, and $\sigma_\tau=0.1$.
These hyperparameters follow the compression--utility operating point identified in prior NVIB
work~\cite{elzein2026differential}.

\paragraph{Efficiency.}
For Llama-3.1-8B, training the Transformer-layer baseline for 5 epochs takes 6.27 h, compared with 7.58 h for NVIB. For Qwen3-8B, training takes 7.13 h and 8.42 h, respectively. Introducing NVIB increases inference latency from 14.30 to 15.68 ms/token (+9.7\%) for Llama-3.1-8B and from 20.30 to 22.12 ms/token (+9.0\%) for Qwen3-8B.

\section{Use of AI Assistants Disclosure}
An AI-based writing assistant was used to improve the grammar and clarity of the manuscript.

\end{document}